%% file: core_main.tex
\documentclass[10pt,twocolumn,letterpaper]{article}
\pdfoutput=1

\makeatletter
\@namedef{ver@everyshi.sty}{}
\makeatother
\usepackage[pagenumbers]{cvpr} 

\usepackage{graphicx}
\usepackage{amsmath}
\usepackage{amssymb}
\usepackage{booktabs}
\usepackage{csvsimple}
\usepackage[shortlabels]{enumitem}

\usepackage[pagebackref,breaklinks,colorlinks]{hyperref}

\usepackage[capitalize]{cleveref}
\crefname{section}{Sec.}{Secs.}
\Crefname{section}{Section}{Sections}
\Crefname{table}{Table}{Tables}
\crefname{table}{Tab.}{Tabs.}

\usepackage{multirow}
\usepackage{xcolor}         
\usepackage{dsfont}
\usepackage{mathrsfs}
\usepackage[group-separator={,}]{siunitx}

\definecolor{CrystalAmethyst}{rgb}{0.6, 0.4, 0.8}
\definecolor{SpiceMelange}{rgb}{0.98, 0.92, 0.84}

\newcommand{\mesh}{$\mathcal{X}^A_b$}
\def\cvprPaperID{11852} 
\def\confName{CVPR}
\def\confYear{2023}

\begin{document}

\title{AnthroNet: Conditional Generation of Humans via Anthropometrics}

\author{Francesco Picetti$^*$
\quad
Shrinath Deshpande$^*$
\quad
Jonathan Leban$^*$
\quad
Soroosh Shahtalebi$^\dagger$
\quad
Jay Patel$^\dagger$
\\
Peifeng Jing
\quad
Chunpu Wang
\quad
Charles Metze III
\quad
Cameron Sun
\\
Cera Laidlaw
\quad
James Warren
\quad
Kathy Huynh
\quad
River Page
\\
Jonathan Hogins
\quad
Adam Crespi
\quad
Sujoy Ganguly$^\S$
\quad
Salehe Erfanian Ebadi$^\S$
\vspace{0.2cm}
\\
\textbf{Unity Technologies}
}


\maketitle

\def\thefootnote{*, $\dagger$}\footnotetext{Equal contribution.}\def\thefootnote{\arabic{footnote}}
\def\thefootnote{$\S$}\footnotetext{Joint corresponding authors.}\def\thefootnote{\arabic{footnote}}

\input{00_abstract}

\section{Introduction}
\label{sec:intro}
\input{01_introduction}
\section{Related Work}
\label{sec:sota}
\input{02_related_work}

\section{Synthetic Data Pipeline}
\label{sec:palette}
\input{03_synthetic_data_pipeline}

\section{Data Preparation}
\label{sec:data}
\input{04_data_preparation}

\section{Methodology}
\label{sec:method}
\input{05_00_methodology}

\section{Experiments and Results}
\label{sec:exp_res}
\input{06_00_experiments_and_results}

\section{Discussion and Remarks}
\label{sec:disc}
\input{07_discussion}

\section{Conclusion}
\label{sec:conclusion}
\input{08_conclusion}


{\small
\bibliographystyle{ieee_fullname} 
\bibliography{core_main}
}

\newpage
\appendix
\label{sec:appendix}
\paragraph{\Large{Appendix}}

\section{Design Parameters}\label{supp:design_parameters}
This section provides a description of all the deep neural networks and the details of the anthropometric measurements introduced in this work.
\input{09_01_mesh_gen_arch}

\input{09_02_anthro_measures_details}

\input{09_03_anthro_regressor}

\input{09_04_mesh_skinner_rigger_arch}

\section{Extended Experiments and Results}\label{supp:extended_experiments}

\input{09_06_extended_results}

\input{09_07_b2a}

\end{document}

%% file: 00_abstract.tex
\begin{abstract}
We present a novel human body model formulated by an extensive set of anthropocentric measurements, which is capable of generating a wide range of human body shapes and poses. The proposed model enables direct modeling of specific human identities through a deep generative architecture, which can produce humans in any arbitrary pose. It is the first of its kind to have been trained end-to-end using only synthetically generated data, which not only provides highly accurate human mesh representations but also allows for precise anthropometry of the body. Moreover, using a highly diverse animation library, we articulated our synthetic humans' body and hands to maximize the diversity of the learnable priors for model training. Our model was trained on a dataset of $100k$ procedurally-generated posed human meshes and their corresponding anthropometric measurements. Our synthetic data generator can be used to generate millions of unique human identities and poses for non-commercial academic research purposes\footnote{\url{https://unity-technologies.github.io/AnthroNet/}}.

\end{abstract}

%% file: 01_introduction.tex

\begin{figure*}[htb!]
    \centering
    \includegraphics[bb=0 0 1920 1080, width=\textwidth, trim={0cm 23.6cm 18cm 0cm}, clip]{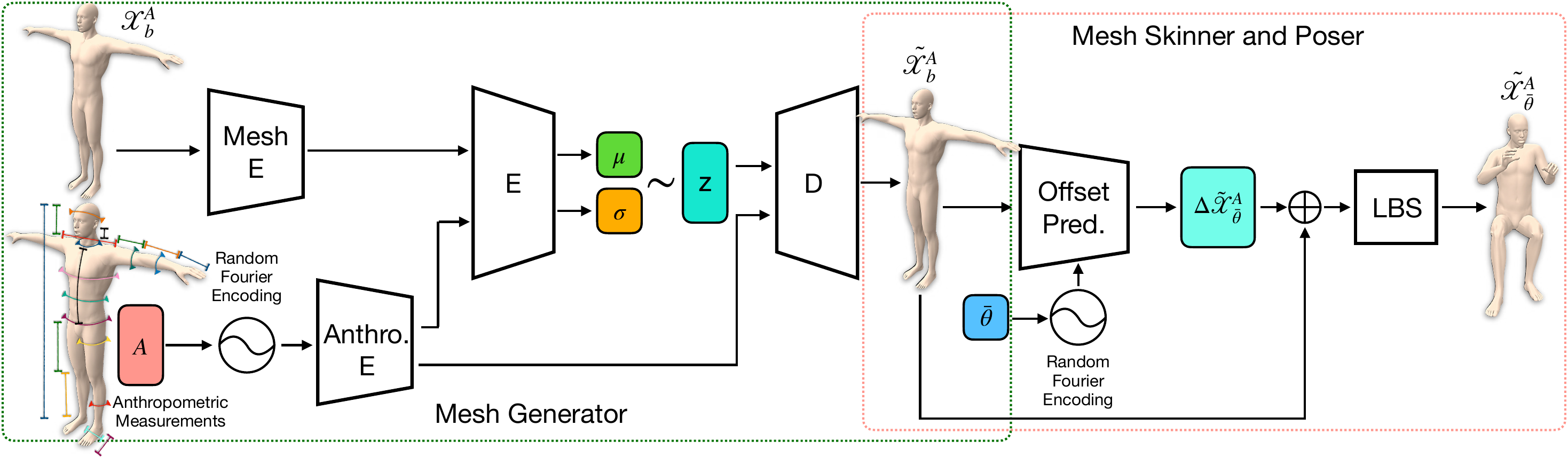}
    \caption{\textbf{AnthroNet's end-to-end trainable pipeline.} 
    We discard the Mesh Encoder in the Mesh Generator block at inference time. Then we decode the estimated mesh in the bind pose $\Tilde{\mathcal{X}}^A_b$A using a latent vector $Z$ sampled from a normal distribution combined with the encoded anthropometric measurements $A$.
    The mesh Skinner and Poser block predicts the pose and shape corrective offsets $\Delta \Tilde{\mathcal{X}}^A_{\bar{\theta}}$ required to animate the mesh into the desired pose $\bar{\theta}$ and produce a fully skinned, rigged, and posed mesh $\Tilde{\mathcal{X}}^A_{\bar{\theta}}$ with the provided measurements $A$.}
    \label{fig:model_arch}
\end{figure*}

\begin{figure*}[htb!]
    \centering
    \includegraphics[bb=0 0 1920 700, width=\textwidth, trim={0cm 0cm 0cm 0cm}, clip]{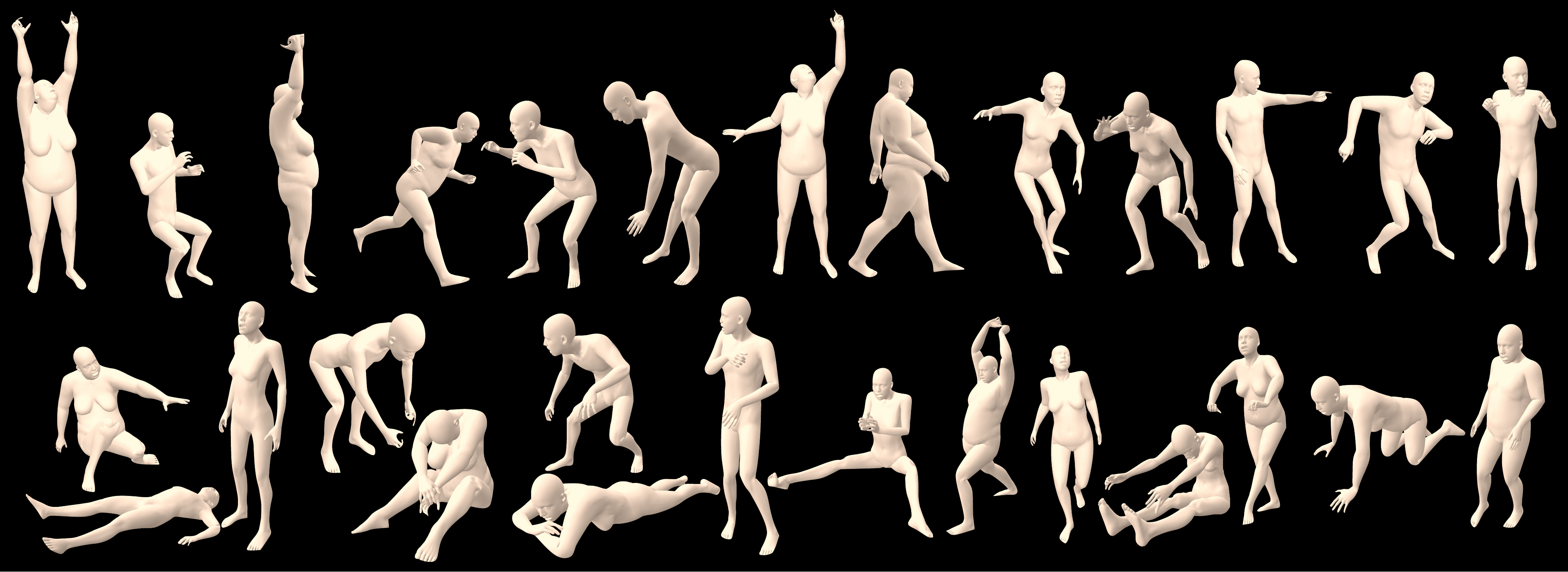}
    \caption{\textbf{Examples from our multi-subject and multi-pose synthetic dataset.}}
    \label{fig:synth_data}
\end{figure*}

With the rise of digital worlds, and the unprecedented need to replicate and represent digital copies of real-world objects, there has been a significant surge of interest in developing methods for representing objects in 3D.
These 3D representations are particularly important for human-centric computer vision and graphics problems like person tracking, human action recognition, and virtual try-on applications. 
Advanced human body models representing the most extensive range of possible human physiques, expressions, and motions are critical for these applications.
Seminal works such as the SMPL(-X) body model~\cite{loper2015smpl, SMPL-X:2019} and the subsequent improvements and extensions~\cite{STAR:2020, FLAME:SiggraphAsia2017, MANO:SIGGRAPHASIA:2017} played a critical role in creating a realistic animated human body model that can represent many different body shapes. 
Later on, GHUM(L)~\cite{xu2020ghum} created an accurate human body model based on a set of deep generative models that each encapsulates variations on the human body, hands, head, and pose. 
One major limitation of the preliminary works on human shape modeling is their reliance on dense body scans that often require expensive motion capture and scan setups, which prohibit data collection and annotation at large scales. 
Thus, there has been a growing interest in developing models where the body shape is regressed from monocular images~\cite{zhang2021pymaf, kocabas2021pare, kocabas2020vibe, kocabas2021spec}. 

Recently, SHAPY~\cite{Shapy:CVPR:2022} presented a new image-based paradigm, where a combination of linguistic shape attributes and some basic anthropometric measurements -- i.e., height, weight, chest circumference, waist circumference, and hips circumference -- to condition the regression of the SMPL-X model parameters from monocular RGB images. 
Although SHAPY outperforms its counterparts in regressing a 3D shape from monocular images, deploying linguistic shape attributes reported subjectively with no ground truth for validation can introduce noise to the model. 
Moreover, their limited set of anthropometric measures only defines the body's overall shape and limits the degrees of freedom needed to represent a wide variety of shapes.  

Inspired by SHAPY, we present a novel non-linear human body model conditioned on an extensive set of objective anthropometric measures that allows us to represent a wide range of human body shapes. Our proposed generative model AnthroNet is trained end-to-end on posed human body meshes (Fig.~\ref{fig:model_arch}) and their associated anthropometric measures. To train AnthroNet, we procedurally generated \num[group-separator={,}]{100000} multi-subject multi-posed human meshes.
This dataset includes the mesh and pose data and the anthropometric measures of the meshes. Finally, we provide an anthropomorphic measurement regressor, which after registration with AnthroNet, allows us to easily extract anthropometric measurements from any real-world scan or other body models without the need to define mesh-specific methods to extract the anthropometric measures carefully.     
In summary, our contributions are:
\begin{itemize}
    \item AnthroNet provides a novel expressive high resolutions model of human body shape conditioned on anthropomorphic measurements.
    \vspace{-0.2cm}
    \item AnthroNet offers two efficient registration pipelines that enable a bi-directional conversion of SMPL and SMPL-X meshes into AnthroNet meshes. 
    \vspace{-0.2cm}
    \item Our synthetic data generator which can produce high resolution meshes with their associated metadata is publicly available for non-commercial academic research use.
\end{itemize}

%% file: 02_related_work.tex
The field of human body modeling has seen a rich body of work on 3D articulated mesh reconstruction for the whole body, including hands and face~\cite{loper2015smpl, SMPL-X:2019, xu2020ghum, allen2003space, anguelov2005scape, hasler2009statistical, katircioglu2018learning, joo2018total, pavlakos2019expressive, loper2014mosh, STAR:2020} and individual body parts alone for a finer level of expressiveness and control~\cite{MANO:SIGGRAPHASIA:2017, FLAME:SiggraphAsia2017, taylor2016efficient, oberweger2015training, ploumpis2019combining}. 
More recently neural parametric models~\cite{palafox2021npms} and implicit parametric models~\cite{palafox2022spams, bozic2021neural, xiu2022icon, alldieck2021imghum} for 3D deformable shapes have been proposed which leverage the flexibility of the learned implicit models that model the body as continuous functions, where the surface is represented as the level set of a neural network that is defined on the 3D space~\cite{park2019deepsdf}. 

Most of these techniques relied on dense point clouds obtained from 3D scans or used an already trained parametric model of the human body to reconstruct the 3D shape from images, videos, motion capture data, or other modalities.
To date, SMPL and SMPL-X remain the most widely used body models for their simplicity and ease of integration~\cite{peng2021neural, muller2021self, Bogo:ECCV:2016, prokudin2021smplpix, AMASS:ICCV:2019, alldieck2018video, kolotouros2019learning, santesteban2020softsmpl, zhi2020texmesh, zhang2017detailed, alldieck2019tex2shape, joo2018total, kocabas2021spec, kocabas2020vibe, kocabas2021pare}. 
More recently, GHUM and GHUML~\cite{xu2020ghum} proposed an end-to-end trainable deep architecture that allows for higher resolution complete body modeling with \num[group-separator={,}]{10168} and \num[group-separator={,}]{3194} mesh vertices respectively. 
However, there is no mechanism by which a specific body shape from the embedding of their trained model can be sampled and reconstructed. 
In contrast, our method allows for such sampling, as the body shape information is used as a conditioning augmentation vector in our generative pipeline. 
Since we trained our model on synthetic data, we opted for the highest resolution mesh representation with \num[group-separator={,}]{36078} mesh vertices, more than $3\times$ that of GHUM. 
This higher-resolution representation allows for more expressive facial, hand, and body shapes. 
Using synthetic data for training means we do not need to rely on mesh registration pipelines from 3D point clouds~\cite{li2019lbs, jiang2019skeleton}. 
Instead, we obtain perfect mesh ground truth, which helps with consistent and error-free modeling of articulations on the mesh.

A large body of work has been dedicated to 3D pose and shape estimation from images and videos~\cite{alldieck2019tex2shape, SMPL-X:2019, zhang2021pymaf, alldieck2018video, kocabas2021pare, Bogo:ECCV:2016, joo2018total, kanazawa2019learning, kocabas2020vibe, kocabas2021spec, kolotouros2019learning}. 
Most techniques rely on SMPL and its variants due to the simplicity of the PCA priors for modeling the human body.
More advanced dimensionality reduction techniques have also been proposed based on Gaussian Mixture Models~\cite{Bogo:ECCV:2016} and Variational Autoencoders (VAEs)~\cite{SMPL-X:2019} that allow for higher accuracy and generalization to more poses. 
GHUM~\cite{xu2020ghum} combined VAEs and normalizing flow representations for the body and facial deformation and the skeleton kinematics, respectively. 
In our technique, we use a single conditional VAE (CVAE) that allows for anatomically consistent and robust pose and shape modeling while enabling the users to define strict priors for the body representation. 
This opens a new avenue for technical artists and other less artistically inclined users to take advantage of a powerful creation tool that can statistically represent high-resolution 3D articulated human body shape and pose in a single learned embedding with fast inference.

In SHAPY~\cite{Shapy:CVPR:2022}, sparse anthropometric measurements and linguistic attributes, \textit{e.g.}, ``short/tall'', ``long legs'', ``pear-shaped'', were used as constraints for 3D shape prediction. 
The sparsity of those measurements and the natural ambiguity that arises from using linguistic attributes to describe a human body shape only provide coarse and global body proportions and embedded correlation in proportionality and shape present in the SMPL training data. 
For example, a ``muscly'' person may produce a person whose entire body is proportionally large. 
In contrast, people with a larger upper body than a lower body can exist.
Our model fills this knowledge gap as it was trained with dense anthropometric measurements covering the whole human body. 
The following sections introduce our synthetic data generation pipeline and describe our human body model.

%% file: 03_synthetic_data_pipeline.tex
Conventionally, models have been trained on collections of registered meshes with pre-defined topologies onto real scans of actors in motion capture studios. 
This process is limited by the number and diversity of people scanned and the set of actions they can perform. 
Additionally, this data suffers from small-scale registration errors.
Some efforts have been made to provide human-understandable textual descriptions for human bodies~\cite{Shapy:CVPR:2022}. 
However, they are still far from providing accurate, consistent, and true-to-life measurements of the subjects. 
Such scans also lack the diversity of human bodies they represent, as clothing models are not representative of the entire population.

Synthetic data can generate millions of unique humans with perfect measurements and annotations, providing more diversity and accuracy compared with their real data counterparts. 
Prior work has demonstrated how diverse and perfectly labeled synthetic data can improve model performance on large benchmark tasks~\cite{ebadi2021peoplesanspeople, ebadi2022psphdri, wood2021fake}. 
Access to the mesh in a game engine enables us to take accurate and consistent anthropometric measurements of our characters, just like an experienced tailor would in reality. 
These accurate anthropometric measurements facilitate training a single model representing all humans, children, and adults, with all body variations and types. 
The same anthropometric measurements used as conditioning augmentation can help us obtain more disentanglement for body representations in the learned manifold of our model, directly resulting in training a much more expressive body model.

A team of technical artists hand-sculpted extreme ends of the body morphology and a few intermediary states to create a procedural pipeline capable of generating such diverse labeled data. 
Heavy and lean characters were hand sculpted in ZBrush for males and females separately while respecting the realistic physical constraints of the human body. 
Then using blend shape logic in Houdini, intermediary states can be obtained for characters of varying weight. 
The same was done for age, height, ethnicity, and sex. 
They were then imported into the Unity game engine as Vertex Animated Textures (VAT), which stores the blendshapes in a storage-friendly texture format for use in the Unity Synthetic Humans package. 
This enables us to sample from the entire set of possible humans randomly. 
As the blend shape logic can vary the mesh vertex positions by infinitesimally small real-valued deltas, our characters' search space is practically infinite. 
The resulting meshes are high-poly representations with \num[group-separator={,}]{36078} mesh vertices and \num[group-separator={,}]{68414} mesh faces (triangles). 
In addition, technical artists hand-drew the appropriate skinning weights for male and female characters, valid across the entire set of produced meshes. 
It can deform well to a large realistic motion capture dataset. We later use these same skinning weights along with the joint-vertex correspondences in our mesh skinning and posing pipeline to re-create a fully rigged, skinned, and posed human character given an unseen input set of anthropometric measurements and desired pose.


%% file: 04_data_preparation.tex
Using our synthetic data generator, by random sampling from a uniform distribution along each axis of variation, e.g., sex, age, height, ethnicity, weight, and facial identity, we generated a total of \num[group-separator={,}]{100000} distinct human body shapes that are used in our training and validation pipeline. 
Each human is generated in bind pose (T-pose) and a random pose, where the latter is randomly sampled from a large dataset of \num[group-separator={,}]{1778} diverse and complex animation clips that are created from real motion capture sequences performed by multiple subjects. 
Since our generated characters' morphology may widely diverge from those of the actors in the motion capture sequences, some self-intersection is inevitable due to the simple re-targeting pipeline we used. 
However, character animation and re-targeting are outside the scope of this work, and the posed characters produce sufficient results for training a human body model.
The Unity Perception package~\cite{borkman2021unity} provides highly accurate 3D keypoint locations and rotations for our characters. 
In our mesh generation pipeline, we rely on 3D vertex locations on the mesh, represented in the local character space with the root joint (hip) at the origin. Furthermore, we use the mesh faces (triangles) in the optimization to enforce surface smoothness.
The 3D keypoint locations (joint locations) in the posed mesh are used in our mesh poser pipeline. 

%% file: 05_00_methodology.tex
Our end-to-end trainable pipeline comprises a conditional variation autoencoder (CVAE) for human body modeling and mesh generation and a deep mesh rigging, skinning, and posing component.
Our mesh generation component produces a human mesh in bind pose,
given a set of anthropometric measurements as conditioning vectors, each of which we represent as Fourier encodings.
The second component takes a mesh in bind pose and produces the vertices offsets relative to the desired pose. 
We obtain the final posed mesh by combining such offsets and the bind pose vertices through a skinning function.


\subsection{Anthropometric Measurements}
\label{subsec:anthro}
\input{05_01_anthropometric_measurements}

\subsection{Random Fourier Encoding}
\label{subsec:fourier}
\input{05_02_fourier}

\subsection{Conditional Human Mesh Generation}
\label{subsec:mesh_gen}
\input{05_03_cvae}

\vspace{0.05cm}
\subsection{Deep Mesh Rigging, Skinning, and Posing}
\label{subsec:skinning}
\input{05_04_skinning}

\vspace{0.05cm}
\subsection{Mesh Registration Pipeline}
\label{subsec:registration}
\input{05_06_mesh_registration}


%% file: 05_01_anthropometric_measurements.tex
Anthropometric measurements consistently and accurately represent information for many human body shapes and sizes.
Extensive studies have been devoted to qualitative and quantitative measurement of human populations~\cite{tsoli2014model, yan2020anthropometric, aslam2017automatic, wuhrer2013estimating, cdc1988anthropometry, gordon20142012, hynvcik2021personalization}.
We defined a dense set of 36 anthropometric measurements over the human body, representing the height and weight of the human, characterized by measurements over the head, neck, shoulders, torso, chest, waist, hips, arms, hands, thighs, legs, and feet. For each measurement, a technical artist has defined a set of vertices over which the measurement is made, such that it consistently represents a tape measure in real life as close as possible. The body part lengths are the euclidean distances between specific vertices on the mesh. 
By registering any 3D human scanned data or human mesh to its closest approximation with our human body model, we can provide the anthropometric measurements of any 3D human scan similarly to~\cite{tsoli2014model}. Since our synthetic human dataset was created with true-to-life scales and proportions, our anthropometric measurements provide close to actual real-life metric body size measurements. Leveraging such information for all existing human scan data is a valuable contribution that we shall leave to future work, due to time constraints. Furthermore, whilst we opted for the 36 anthropometric measurements over our meshes -- which provide the maximal meaningful information about all our meshes -- obtaining more measurements on every edge loop on the synthetic meshes is a straightforward task.

%% file: 05_02_fourier.tex

Fourier encoding~\cite{tancik2020fourfeat} is an approach that maps continuous input coordinates into higher dimensional space to help recover high-frequency details from the input data. 
We deal with 3D mesh representations encapsulating position-dependent information; further, the exact anthropometric measurements over body regions could come from two slightly different human body shapes; also, each measure has different scales; therefore, the Fourier encoding of such input data seems to be a natural choice. For us, this directly translated to much higher fidelity in the preservation of geometry and further disentanglement of multiple highly correlated anthropometric measurements from one another.

We apply our Fourier encoding on our anthropometric measurements and the pose input $\theta$ to the network in the form of global locations of body joints.
For each input point $y$ and given a random Gaussian matrix $B$, whose entries are drawn independently from a normal distribution $\mathcal{N}(0, \sigma^2)$ we encode our input as:
\begin{equation}
    \mathscr{F}(y) = [\cos(2 \pi B y) \quad \sin(2 \pi B y)]^T,
\end{equation}
which is a deterministic mapping and can be computed feasibly before training.

%% file: 05_03_cvae.tex
We formulate the human mesh generation objective as learning the conditional distribution of observed human meshes with respect to their corresponding anthropometric measurements.
We employ a conditional latent variable generative model. 
The high-level architecture of this model was based on Conditional Variational Autoencoder (CVAE)~\cite{cvae} as shown in Fig.~\ref{fig:model_arch}.
Forward pass of the CVAE architecture starts with an encoder that takes in the input as a tuple of vectors, namely human mesh in bind pose \mesh, and a vector of conditionals $A$ which are the Fourier encoded anthropometric measurements.
Each of the vectors are passed through separate sub-encoding modules which are fused later to produce a probability distribution $p(z)$ for embedding, similar to VAE.
A sample is taken from this probability distribution, along with encoded conditional vector at the decoder stage to produce an output mesh $\Tilde{\mathcal{X}}^A_b$.
We train this generative model on the total loss given by
\begin{equation}\label{eq:eq_cvae}
    \mathcal{L}_\text{G} = \alpha \, \mathcal{L}_{\text{rec}}\left(\mathcal{X}^A_b, \Tilde{\mathcal{X}}^A_b\right)\\
    + \beta \,  \mathcal{L}_{\text{KL}}(z)\\
    + \gamma \,  \mathcal{L}_{\text{Lapl}}\left(\Tilde{\mathcal{X}}^A_b\right),
\end{equation}
where $\mathcal{L}_{\text{rec}}$ and $\mathcal{L}_{\text{KL}}$ are the reconstruction and Kullback-Leibler divergence losses similarly to the VAE loss.
Specifically, $\mathcal{L}_{\text{rec}}$ is the Huber distance between the generated mesh $\Tilde{\mathcal{X}}^A_b$ and the target mesh $\mathcal{X}^A_b$:
\begin{equation}\label{eq:huber}
    \mathcal{L}_{\text{rec}} =
    \begin{cases}\frac{1}{2}\left(\mathcal{X}^A_b-\Tilde{\mathcal{X}}^A_b\right)^2 & \text { if }\left|\mathcal{X}^A_b-\Tilde{\mathcal{X}}^A_b\right| \leq  1 \\  \left|\mathcal{X}^A_b-\Tilde{\mathcal{X}}^A_b\right|-\frac{1}{2} & \text { otherwise. }\end{cases}
\end{equation}
$\mathcal{L}_{\text{KL}}$ quantifies the divergence between the latent space distribution $p(z)$ and a reference distribution $~\mathcal{N}(0,1)$:
\begin{equation}\label{eq:kl}
    \mathcal{L}_{\text{KL}} =  \sum_{z \in \mathcal{Z}} p(z) \log \frac{p(z)}{\mathcal{N}(0,1)}.
\end{equation}
We include a Laplacian loss that penalizes high frequency variations on the reconstructed mesh surfaces:
\begin{equation}\label{eq:laplacian}
    \mathcal{L}_{\text{Lapl}}\left(\Tilde{\mathcal{X}}^A_b\right) =
    \sum_{v=1}^V
    \sum_{n \in \mathcal{R}(v)}
    \Vert \ell_{v,n} \left(\Tilde{x}_v^A - \Tilde{x}_n^A\right)\Vert_2^2,
\end{equation}
where $V$ is the number of vertices in the mesh, the ring $\mathcal{R}(v)$ is the set of neighboring vertices to the $v$-th vertex and $\ell_{v,n}$ are cotangent-based Laplacian weights.

We performed ablation experiments where random Fourier encoding on the conditionals before passing them to the encoder yielded the best results on our test set.
During inference, the mesh encoder is discarded, and we only use decoder side of the model, which takes in the encoded anthropometric information and a latent vector and produces a sample from the conditional distribution given by
\begin{equation}\label{eq:generation}
     \Tilde{\mathcal{X}}^A_b = D (A, z).
\end{equation}


%% file: 05_04_skinning.tex
Many popular human body models~\cite{loper2015smpl, xu2020ghum} and 3D Digital Content Creation (DCC) applications use some form of Linear Blend Skinning (LBS) or Dual Quaternion Skinning in order to bind the mesh to the skeleton and animate the mesh along with the skeleton. We opted to create a body model that is compatible with other graphics software and can be used as a drop-in placement in applications that use the same skinning pipeline. 
In our end-to-end trainable pipeline, we use a variant of LBS, where the skinning weights $\omega_{v,j}$ for each vertex $v$ and joint $j$, are provided to us from the DCC application and are consistent across all body sizes and poses. In particular, the skinning weights are differently computed for male and female meshes.

For each character, we use the global joint locations $\theta$ from the bind pose and the global joint locations $\Bar{\theta}$ from the current arbitrary pose which are calculated from the global mesh vertex positions in the bind pose and current pose, respectively. Then given the pose and shape-dependent corrective offsets $\Delta \Tilde{\mathcal{X}}^A_{\theta}$ that are learned in our deep pose corrective module, we are able to skin the mesh using the modified LBS formulation. As such, each vertex on the posed mesh can be formulated as follows in homogeneous coordinate system:

\begin{equation}
    x^A_v = \omega_{v,j} T_j(\theta) T_j(\Bar{\theta})^{-1}
    \begin{bmatrix}
        \Tilde{x}^A_v + \Delta \Tilde{x}^A_{v, \theta}\\
        1
    \end{bmatrix}
\end{equation}

\begin{equation}
    T_j(\theta) = 
    \begin{bmatrix}
        \mathds{1} & \theta\\
        0 & 1
    \end{bmatrix} \in SE(3),
\end{equation}
where $T_j(\theta)$ is the global transformation matrix for joint $j$. Then we compute the transformation from bind pose to posed mesh by multiplying the inverse of the global transformation matrix at bind pose $\theta$ and the global transformation matrix at current pose $\Bar{\theta}$.

Prior works~\cite{loper2015smpl, xu2020ghum} have relied on learning pose-corrective and shape-corrective blend shapes separately. In contrast, we learn both the pose and shape-dependent corrective offsets in a single model. This allows us to learn multi-subject and multi-pose dependent corrective offsets in the hidden space of our mesh skinner regression model. The mesh skinner model takes as input the predicted bind pose mesh $\Tilde{\mathcal{X}}^A_b$ from our mesh generator, given the anthropometric measurements $A$. It then fuses them with the current pose $\mathscr{F}(\bar{\theta})$ which are encoded with our random Fourier encoding $\mathscr{F}$. The mesh skinner is then asked to predict the pose and shape-dependent corrective offsets $\Delta \Tilde{\mathcal{X}}^A_{\bar{\theta}}$ that can produce the skinned and posed mesh $\Tilde{\mathcal{X}}^A_{\bar{\theta}}$ in the LBS formulation above.
To this end, the loss function which the skinner is trained on is
\begin{equation}
  \begin{split}
    \mathcal{L}_\text{S} =
    & \rho \, \mathcal{L}_\text{rec}\left( \Delta \mathcal{X}^A_{\bar{\theta}}, \Delta \Tilde{\mathcal{X}}^A_{\bar{\theta}} \right) + \\ 
    & \eta \, \mathcal{L}_{\text{rec}}\left(\mathcal{X}^A_{\bar{\theta}},  \Tilde{\mathcal{X}}^A_{\bar{\theta}} \right) +
    \lambda \, \mathcal{L}_{\text{Lapl}}\left(\Tilde{\mathcal{X}}^A_{\bar{\theta}} \right),
  \end{split}
\end{equation}
where the first term is the Huber distance between the desired offsets and the network output; the second term is the Huber distance between the desired and reconstructed posed meshes;
and the third term constrains the optimization to obtain smooth meshes.

%% file: 05_06_mesh_registration.tex
Mesh registration is an optimization routine that we use at different stages of our pipeline in order to convert body meshes from different models to each other. Due to the popularity of SMPL-X, we developed the following two pipelines.
The reader is referred to the appendix for further details on the qualitative results.
\vspace{-0.4cm}
\paragraph{AnthroNet-to-SMPLX}
Here the goal is to build an AnthroNet mesh that fits a given SMPL-X mesh, so as to compare the expressiveness of AnthroNet to SMPL-X.
We define $c$ and $\theta$ as the learnable variables which will be updated through a gradient-based optimization. Here $c \in \mathbb{R}^{37}$ represents concatenation of sex with the list of anthropometric measures, and $\theta$ denotes the vector of joint locations. The mesh generator and the pose module, both with frozen parameters, take $c$ and $\theta$ as inputs. We minimize the Chamfer loss between the generated posed mesh and the target SMPL-X mesh, and the gradients are calculated accordingly to update the learnable parameters.  
\vspace{-0.4cm}
\paragraph{SMPLX-to-AnthroNet}
This pipeline follows a similar logic as in AnthroNet-to-SMPLX but with the opposite goal of converting the SMPL-X meshes to AnthroNet. The major difference here is that prior to calculating the Chamfer loss over the vertices, we first minimize the loss on the joint locations until it reaches a certain range; afterwards the Chamfer loss comes into effect for the optimization process.

%% file: 06_00_experiments_and_results.tex

\begin{figure}[!htb]
    \centering
    \begin{subfigure}[t]{\linewidth}
        \includegraphics[bb=0 0 747 399, width=\linewidth, trim={0.5cm 0.5cm 0.5cm 0.5cm}, clip]{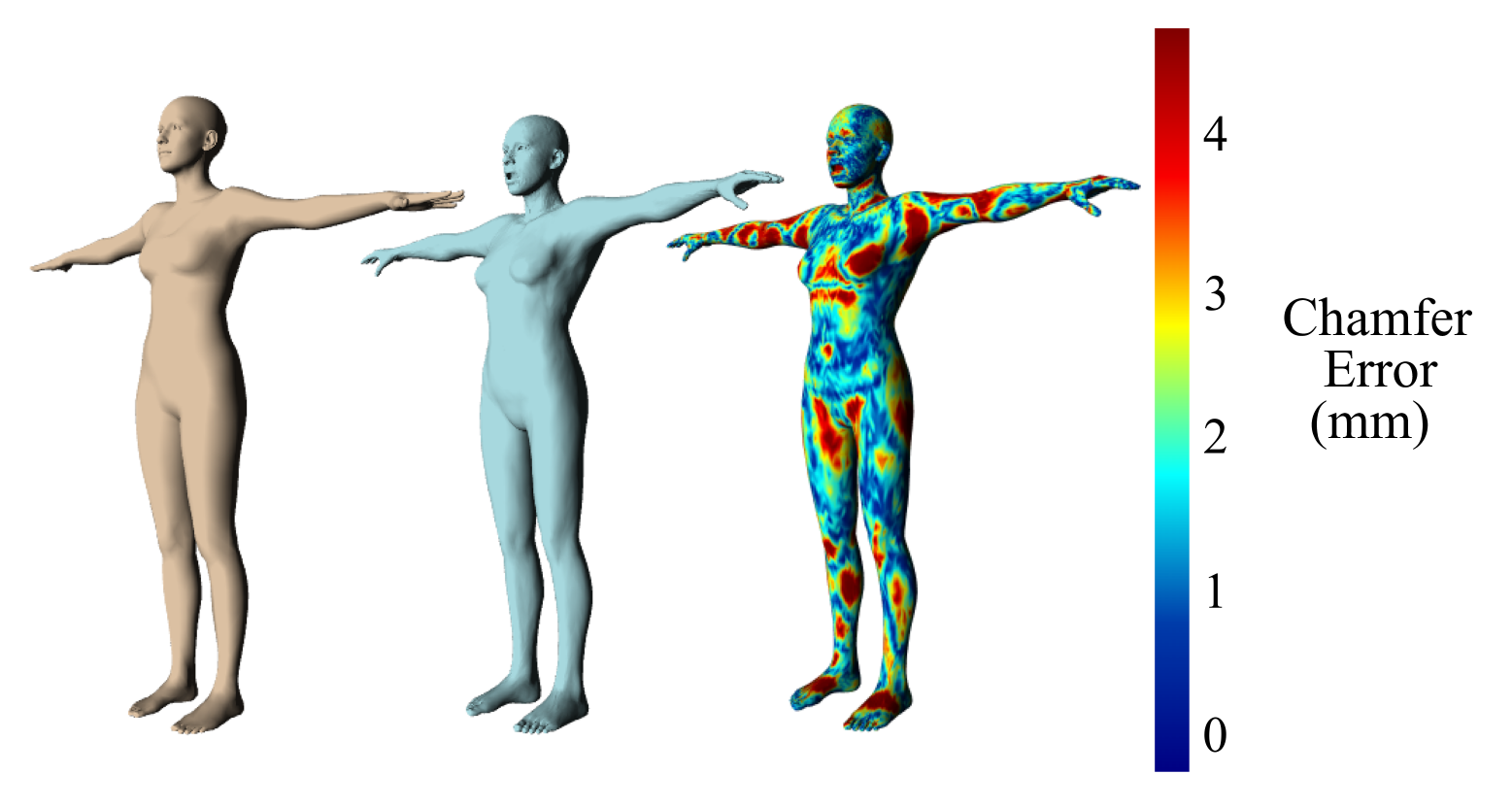}
        \label{fig:buffers-mask}
    \end{subfigure}
    \vspace{-2em}
  \caption{\textbf{Illustration of Registration Error when AnthroNet is fitted to a SMPL-X mesh.} A large portion of error attributes are due to the fact that SMPL-X meshes are registered on humans wearing tight clothing, whereas our synthetic humans are without clothes.}
  \label{fig:fig_registration_error}
\end{figure}

\begin{figure}[!htb]
  \centering
  \begin{subfigure}[t]{\linewidth}
        \includegraphics[bb=0 0 517 237, width=\linewidth, trim={1cm 0cm 0cm 0.5cm}, clip]{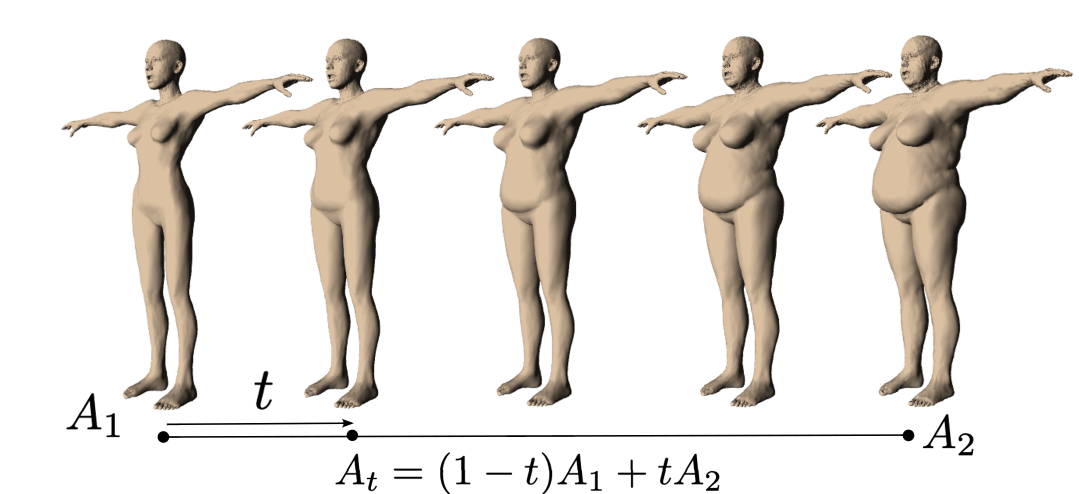}
        \caption{Lean to curvy; keeping the same height and sex.}
        \label{fig:buffers-mask}
    \end{subfigure}
    \hspace{1.5em}
    \begin{subfigure}[t]{\linewidth}
        \includegraphics[bb=0 0 487 214, width=\linewidth, trim={0.5cm 0.5cm 0.5cm 0.5cm}, clip]{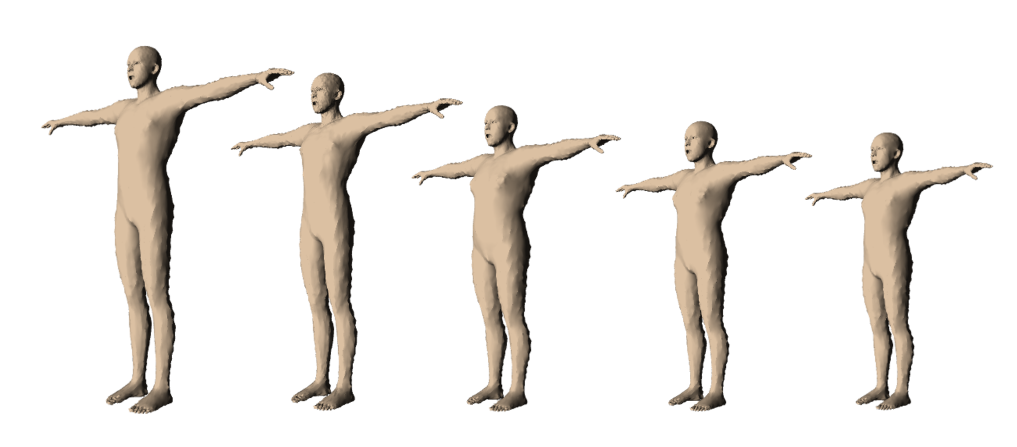}
        \caption{Tall to short; same sex and chest-to-waist circumference ratio.}
        \label{fig:buffers-nopb}
    \end{subfigure}
    \begin{subfigure}[t]{\linewidth}
        \includegraphics[bb=0 0 508 215, width=\linewidth, trim={0.5cm 0.5cm 0cm 0.5cm}, clip]{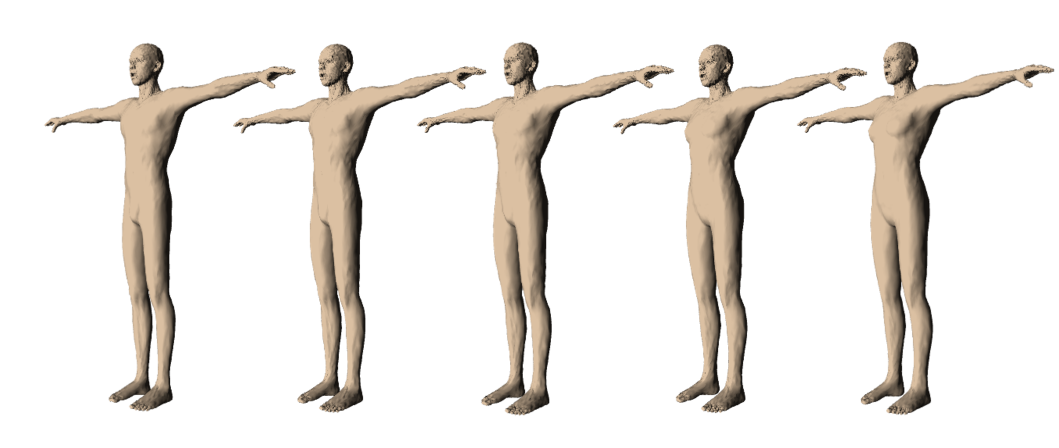}
        \caption{Male to female; keeping the same anthropometric measurements.}
        \label{fig:buffers-nopb}
    \end{subfigure}
  \caption{\textbf{Human meshes generated by AnthroNet conditioned on measurements and sex.} As depicted in (a) a mesh is generated using conditional $A_{t}$, which is obtained from the linear interpolation between two measurement vectors ($A_1$, $A_2$).}
  \label{fig:interpolation}
\end{figure}

We evaluate the performance of AnthroNet as a whole and its individual modules, namely Mesh Generator, Poser, and the Registration module. We have also performed an extensive set of supplementary experiments and ablations which are provided in the appendix. Throughout this section, the Mean Point-to-Point Error (P2P)~\cite{Shapy:CVPR:2022} is reported as the error between two meshes. 
\vspace{-0.4cm}
\paragraph{Generative Performance} The Mesh Generator is tasked with producing a T-posed human mesh that corresponds with a set of anthropometric measurements. 
We first evaluate this module's performance on our held out test set produced by our synthetic data generator.
We use this criteria to determine the best conditional mesh generator on the synthetic data.
This was done by conducting ablation studies to understand the impact of different components in the compute graph (for details see appendix). 
The ablation studies indicated that random Fourier encoding lead to better conditional generation as summarized in Tab.~\ref{tab:generative_ablation_table}. 
\vspace{-0.4cm}
\paragraph{Mesh Registration Performance} The SMPL-X body model is widely used in the literature; thus we perform registration on 1185 SMPL-X meshes obtained from SHAPY fashion model dataset~\cite{Shapy:CVPR:2022}. Since our human mesh contains significantly more vertices, we re-sample vertices on the SMPL-X mesh such that they are equal to the vertices in our model. 

Tab.~\ref{tab:registration_error} presents statistics of registration results, which indicate a good overall fit based on Euclidean Chamfer distance between our model's generated mesh and the re-sampled SMPL-X mesh target. As it can be seen, there is a high amount of error in the chest area particularly for female meshes; this is due to the fact that scans of female subjects used to train the SMPL-X model were wearing tight clothing and sports bras.
Fig.~\ref{fig:fig_registration_error} depicts visual error analysis on the registration, highlighting zones with topological differences in the modeling.

\begin{figure*}[htb!]
    \centering
    \includegraphics[bb=0 0 795 270, width=\textwidth, trim={1.5cm 0.5cm 1.5cm 0.5cm}, clip]{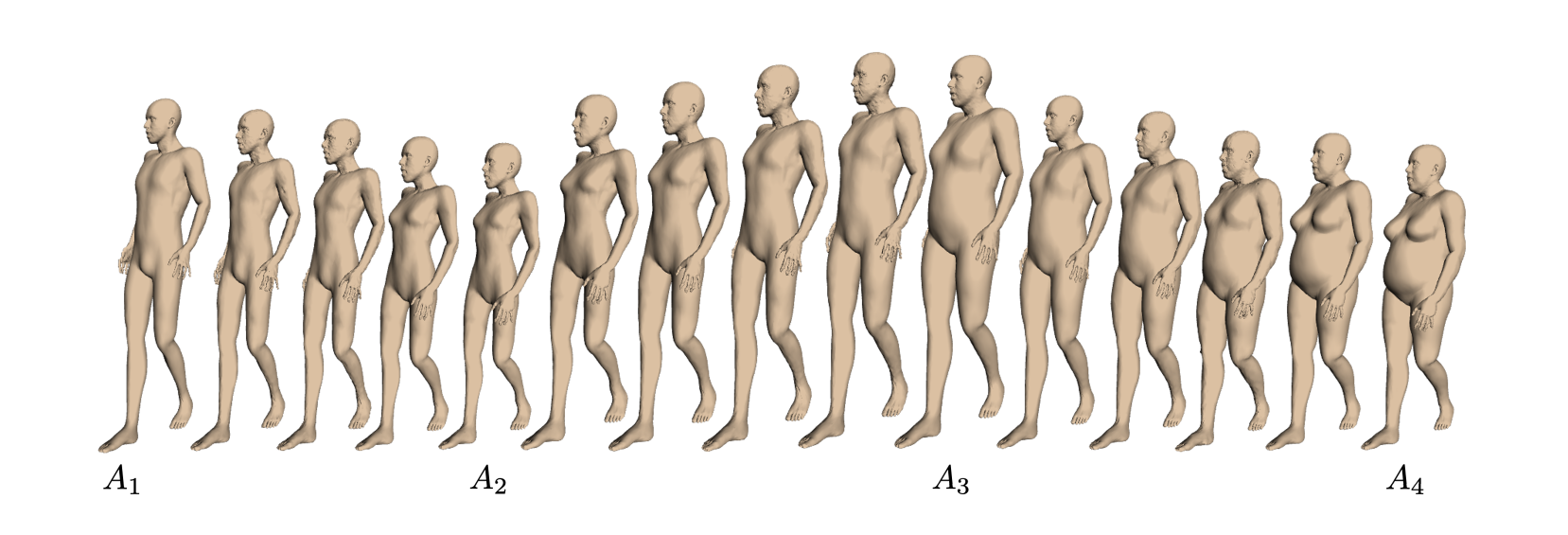}
    \caption{\textbf{AnthroNet with Poses}. The figure shows posed meshes obtained by applying LBS to generated meshes on conditionals computed by piece-wise linear interpolation between four sets of full anthropometric measurements ($A_1, A_2, A_3, A_4$); for details see the appendix.}
    \label{fig:interpolation_posed}
\end{figure*}

\vspace{-0.4cm}
\paragraph{Interpretability of Conditionals} Having trained with precise anthropometric data as conditionals, AnthroNet has interpretable and precise control over generation of human meshes.
We demonstrate this ability in Fig.~\ref{fig:interpolation}, where only certain anthropometric measurements are varied to generate a desired human shape using equation~(\ref{eq:generation}).
The meshes shown are computed by taking two conditional points in 37 dimensional space (36 measurements plus sex). 
The sample points are selected such that they describe the desired qualitative aspects of a person; e.g., a lean female of a certain height.
Then a measurement vector is computed as the linear interpolation between the two sets. 
Fig.~\ref{fig:interpolation_posed} shows more smooth interpolations in our learned shape manifold where the same pose is applied on all generated meshes.
Here, the conditional vector $A$ is obtained as linear interpolation between conditionals ($A_1, A_2, A_3, A_4$). 
\vspace{-0.3cm}
\paragraph{Regressing Body Shape from Monocular Images} To compare the performance of the proposed AnthroNet with its counterparts on regressing the 3D shape from monocular images, we employ the Human Bodies in the Wild (HBW) dataset~\cite{Shapy:CVPR:2022}. HBW contains 3D body scans of $35$ subjects ($15$ male and $20$ female) along with their registered SMPL-X body models and a total of $2543$ photos of the subjects.
We deploy the image-regression module in SHAPY as a pretrained model with frozen parameters. The goal is to register AnthronNet meshes to SMPL-X meshes with high fidelity and then evaluate the accuracy of the measured anthropometrics on the SMPL-X mesh. 

The image-regression module takes RGB images as input and outputs an estimate of the SMPL-X shape parameters.
To lay a baseline for our comparisons, we apply the measurement module of the SHAPY framework to estimate a sparse set of anthropometric measurements based on predicted SMPL-X parameters. Afterwards, these parameters are passed through our SMPLX-to-AnthroNet pipeline and obtain the corresponding AnthroNet meshes, where our dense set of anthropometric measurements can be calculated.
Finally, we find the correspondence between the dense and sparse measurements and calculate the root mean squared error between the anthropometric measurements and mean vertex error in millimeters.
The results of this experiment are reported in Tab.~\ref{tab:image_regressor_compare}. It is worth highlighting that the deployment of a pretrained image regressor would most likely hinder the whole AnthroNet pipeline to achieve its peak performance due to the potential mismatch between the distributions of data across the two training domains. However, thanks to the registration pipeline, AnthroNet manages to convert SMPL-X models to AnthroNet native mesh system with negligible loss, such that comparable performance on predicting the set of sparse anthropometric measures over HBW dataset is obtained. 


\begin{table}[]
\centering
\resizebox{\columnwidth}{!}{
\begin{tabular}{|l|cccc|}
\hline
\multicolumn{1}{|l|}{Hyper Parameters}                                                             & \multicolumn{1}{c}{P2P} & \multicolumn{1}{c}{Chest} & \multicolumn{1}{c}{Hip} & Waist          \\ \hline
\textbf{\begin{tabular}[c]{@{}l@{}}Full Anthropometrics,\\ Sex, $\gamma = 0$\end{tabular}}       & \textbf{3.83}                                                            & \textbf{46.9}                                                           & \textbf{9.79}                                                          & \textbf{12.3}                                       \\ \hline
\begin{tabular}[c]{@{}l@{}}Full Anthropometrics, \\ Sex, $\gamma = 0.01$\end{tabular}            & 5.55                                                                    & 90.7                                                                     & 38.7                                                                  & 17.6                                               \\ \hline
\begin{tabular}[c]{@{}l@{}}Height, Weight, \\ Sex, $\gamma = 0.01$\end{tabular}                 & 6.55                                                                    & 79.4                                                                    & 27.9                                                                  & 21.3                                               \\ \hline
\begin{tabular}[c]{@{}l@{}}Height, Sex,\\ Select Circumferences, \\ $\gamma = 0.005$\end{tabular} & 6.75                                                                    & 41.8                                                                     & 14.2                                                                  & 13.4                                               \\ \hline
\end{tabular}
}
\caption{\textbf{High-level overview of key ablations.} Fourier Encoding of the conditionals outperforms models without them when tested on held out samples. Models with all anthropometric measurements outperform a subset of select anthropometric measurements. $\gamma$ is Laplacian loss factor as shown in equation~\ref{eq:eq_cvae}.}
\label{tab:generative_ablation_table}
\label{tab:generative_ablation_table}
\end{table}

\begin{table}[]
\centering
\begin{tabular}{|l|c|}
\hline
Body Section & P2P Chamfer distance [mm] \\ \hline
Full Body            & 4.63 $ \pm $ 0.07    \\ 
Chest                & 6.67 $ \pm $ 0.42    \\ 
Waist                & 5.06 $ \pm $ 0.27    \\ 
Hip                  & 6.03 $ \pm $ 0.34    \\ 
Thighs               & 5.15 $ \pm $ 0.16    \\ 
Calves               & 4.67 $ \pm $ 0.08    \\ 
Arms                 & 7.75 $ \pm $ 0.07    \\ \hline
\end{tabular}
\caption{\textbf{Registration error of fitting AnthroNet on 1185 SMPL-X Meshes obtained from Fashion Model Dataset\cite{Shapy:CVPR:2022}.}}
\label{tab:registration_error}
\end{table}

\begin{table}[]
    \centering
    \resizebox{\columnwidth}{!}{%
    \begin{tabular}{|l|c|c|c|c|c|c|}
    \hline
    Method      & Model &  Height  &  Chest &  Waist  &  Hips &  P2P \\
    \hline
    SMPLR~\cite{SMPLR}                              & SMPL   & 182 & 267 & 309 & 305 & 69 \\
    STRAPS~\cite{STRAPS}                            & SMPL   & 135 & 167 & 145 & 102 & 47 \\
    SPIN~\cite{SPIN}                                & SMPL   & 59  & 92  & 78  & 101 & 29 \\
    TUCH~\cite{TUCH}                                & SMPL   & 58  & 89  & 75  & 57  & 26 \\
    Sengupta et al.~\cite{sengupta2021hierarchical} & SMPL   & 82  & 133 & 107 & 63  & 32 \\
    ExPose~\cite{EXPOSE}                            & SMPL-X & 85  & 99  & 92  & 94  & 35 \\
    SHAPY~\cite{Shapy:CVPR:2022}                    & SMPL-X & 51  & 65  & 69  & 57  & 21 \\
    AnthroNet (ours)                                & SMPL-X & $51\pm47 $  & $65\pm16$  & $69\pm22$ & $57\pm16$ & $21\pm4.5$ \\
    \hline 
    \end{tabular}
    }
    \caption{\textbf{Performance comparison of AnthroNet when coupled with a pretrained image regressor.} Values are the error in millimeters and evaluated over the HBW dataset~\cite{Shapy:CVPR:2022}.}
    \label{tab:image_regressor_compare}
\end{table}

%% file: 07_discussion.tex
\paragraph{Limitations}
While our synthetic data is created with rigor and supported by years of experience of technical character artists, it is not free from unintentional errors and biases introduced during the character sculpting, character animation, and retargeting in our animation pipeline, and motion capture animation data that may not represent all possible human motions. We know that the set of identities and ethnicities our synthetic data generator produces does not represent the entire human population. However, we aimed to do due diligence in ensuring that we are not introducing significant biases by over-representing or under-representing specific body shapes, sexes, ethnicities, or ages; hence, we opted for random uniform sampling across the entire set of available parameters for representing a human body.

We measured our anthropometric descriptors over regions of the body that professional tailors use for making garments. Numerous studies on anthropometric measurements~\cite{tsoli2014model, yan2020anthropometric, aslam2017automatic, wuhrer2013estimating, cdc1988anthropometry, gordon20142012, hynvcik2021personalization} suggest that these set of anthropometric measurements best describe variations seen on the human body shape if additional anthropometric measurements are needed, our synthetic data pipeline provisions for such modifications. 
Other human body models are generally trained on 3D scans of individuals who self-identified as male or female or exhibited physical characteristics of one or the other sex. One approach to avoiding limitations arising from over or under-representation of either sex is to train a joint body model on data from both sexes and create a ``neutral'' body model. While this approach has its benefits, we could not do so, as we made the skinning weight data for males and females separately. We shall leave such improvements as future work.

Our generative model aims to learn low-dimensional representations that encapsulate the statistics of the given input data. Its ability is limited by the input data provided and, secondly, by the optimization and metrics used in the learning process. Inevitably, better architectural or model parameter choices and loss function designs will exist that may or may not yield more accurate human shape manifolds.
\vspace{-0.4cm}
\paragraph{Societal Impacts and Ethical Considerations}
Capturing, processing, storing, and transferring any real human data is subject to strict data and privacy regulations and guidelines across the world~\cite{voigt_eu_2017, bukaty_california_2019}. On the other hand, synthetic data allows for preserving privacy and safety and avoiding ethical concerns and biases that may exist with real human data. We sought to provide an alternative to using real human data that addresses some of these concerns. We describe in detail how we have created our synthetic data and body model, and with the provided data, other researchers can reproduce and validate the results.

We make our synthetic data generator publicly available for non-commercial use and academic research. The authors created and vetted the synthetic and motion capture dataset, free from offensive and explicit content and personally identifiable information. We do not anticipate any apparent societal or ethical harm from our work. We believe that relieving the computer vision community of the burden of collecting and annotating vast amounts of human data -- that may bear serious privacy, safety, and ethical concerns -- delivers a net positive impact.
We aimed to provide and pave the path for body models that can better describe attributes relating to the human body shape. We did so by giving dense measurements over the meshes. We hope that this will enable a more inclusive and more expressive model. We also allow the researchers to increase upon the diversity of the synthetic humans by open-sourcing our synthetic data generator.

%% file: 08_conclusion.tex
In this work, we presented AnthroNet, a deep generative human body model that takes an extensive set of anthropometric measurements as input and generates body meshes that are more than three times higher resolution than SMPL-X meshes, with high fidelity to anthropometric descriptions of a human body.
The whole pipeline is trained end-to-end solely on $100,000$ synthetically generated human body meshes,
which not only provide a wide variety of shapes and poses but also alleviate the privacy concerns as well as data collection costs with 3D scanning of real humans.
Lastly, we provide AnthroNet with two registration modules that allow for the bi-directional transformation of SMPL-X meshes to the AnthroNet, which enables obtaining dense anthropometric measurements on all 3D human body scans that can be registered onto the SMPL-X model.

%% file: 09_01_mesh_gen_arch.tex
\subsection{Mesh Generator architecture}
\label{Section:details-mesh-gen}
In this section, we describe the design details of the deep neural network responsible for the mesh generation.
In this conditional Variational Auto-Encoder (cVAE), we can identify four main components:
\begin{enumerate}[i.]
    \item a mesh encoder takes the vertices target mesh in bind pose $\mathcal{X}^A_b$, and produces a feature vector $h_v$;
    \item a condition encoder takes the anthropometric measurements $A$ (after Fourier encoding) and produces a feature vector $h_a$:
    \item a fusion encoder combines $h_v$ and $h_a$ to get the latent representations $\mu$ and $\sigma$;
    \item a decoder fuses the reparameterized latent space $z$ and the condition feature vector $h_a$ and reconstructs the mesh vertices $\Tilde{\mathcal{X}}^A_b$.
\end{enumerate}

All the encoders share the same core block, composed of a fully connected (FC) layer followed by a batch normalization (BN) layer and a parametric rectified linear unit (PReLU) activation function:
\begin{itemize}
    \item[-] The mesh encoder squeezes the input $3V$ coordinates to $128$ features.
    \item[-] The condition encoder takes $37$ (36 anthropometric measurements and 1 sex identifier) $\times 2f$ ($f$ being the number of Fourier components) inputs and produces $128$ features.
    \item[-] The fusion module takes $[h_v, h_a]^T$ ($256$ features) and passes them in the concatenation of two core blocks. The output of this encoding scheme is a $64$ features vector that is finally processed by two separate fully connected layers to produce $\mu$ and $\sigma$ whose output dimension is $10$.
    \item[-] The decoder is the concatenation of three core blocks; the first one takes $[z, h_a]^T$ and produces $64$ features; the second and the third scales the features up to $128$ and $256$, respectively. Finally, we include an additional FC layer to produce $3V$ coordinates. 
\end{itemize}

%% file: 09_02_anthro_measures_details.tex
\subsection{Anthropometric Measurements}
As mentioned in the core paper, we use 36 measures for the anthropometry of human body. Those measurements are listed in Tab.~\ref{table:statistics-anthro} along with the mean, the standard deviation, the min and max of each measure from our $100k$ dataset.
To lay a baseline, we have reported the statistics of such anthropometric measurements obtained from $7435$ male and $3922$ female subjects, as reported in~\cite{gordon20142012}.
Empty cells denote the absence of a true correspondence between the anthropometric measurement in our dataset and the real measurements.
It is observed that the standard deviation of $34$ out of $36$ anthropometric measurements of our $100k$ dataset is higher than the one of real measurements which underlines on the diversity of our dataset.
 
\begin{table*}[!h]
\centering
\resizebox{\textwidth}{!}{
    \begin{tabular}{|l|c|c|c|c|c|c|c|c|c|c|c|c|}
    \hline
    
    Name & \multicolumn{4}{|c|}{100k Synthetic} & \multicolumn{8}{|c|}{Real~\cite{gordon20142012}} \\ \cline{6-13}
    & \multicolumn{4}{|c|}{} & \multicolumn{4}{|c|}{Male} & \multicolumn{4}{|c|}{Female} \\
    \cline{2-13} 
                            & mean  & std    & min  & max   & mean  & std  & min  & max  & mean  & std  & min  & max   \\ \hline
    Waist circumference     & 0.828 & \textbf{0.174}  & 0.55 & 1.250 & 0.941 & 0.112& 0.648& 1.379& 0.861 & 0.100& 0.611& 1.334 \\
    Chest circumference     & 1.061 & \textbf{0.159}  & 0.76 & 1.522 & 1.059 & 0.087& 0.774& 1.469& 0.947 & 0.083& 0.695& 1.266 \\
    Hip circumference       & 0.988 & \textbf{0.154}  & 0.72 & 1.368 & 0.102 & 0.077& 0.737& 1.305& 1.021 & 0.076& 0.798& 1.341 \\ 
    Height                  & 1.760 & \textbf{0.144}  & 1.49 & 1.991 & 1.756 & 0.0686& 1.491& 1.993& 1.629 & 0.0642& 1.410& 1.829 \\ 
    Shoulder width          & 0.295 & 0.012 & 0.248 & 0.335 & 0.416 & \textbf{0.0192} & 0.337& 0.489& 0.365 & \textbf{0.0183} & 0.283& 0.422\\  
    Torso height from back  & 0.376 & \textbf{0.030} & 0.310 & 0.425 & 0.382 & 0.0259& 0.293& 0.483& 0.350 & 0.0223& 0.284& 0.435 \\ 
    Torso height from front & 0.324 & \textbf{0.029} & 0.256 & 0.394 & 0.478 & 0.027& 0.383& 0.598& 0.425 & 0.026& 0.345& 0.532 \\ 
    Head circumference      & 0.546 & 0.009 & 0.521 & 0.554 & 0.574 & \textbf{0.016} & 0.516& 0.633& 0.561 & \textbf{0.019} & 0.500& 0.635 \\
    Neck circumference      & 0.431 & \textbf{0.044} & 0.320 & 0.541 & 0.398 & 0.026& 0.311& 0.514& 0.330 & 0.019& 0.275& 0.424 \\
    Head height             & 0.245 & \textbf{0.004} & 0.232 & 0.251 & -     & -    & -    & -    & -     & -    & -    & -     \\
    Mid-line neck length     & 0.092 & \textbf{0.013} & 0.056 & 0.115& -     & -    & -    & -    & -     & -    & -    & -     \\
    Lateral neck length     & 0.120 & \textbf{0.012} & 0.078 & 0.141 & 0.108 & 0.012& 0.062& 0.151& 0.106 & 0.012& 0.071& 0.150 \\ 
    Hand size               & 0.177 & \textbf{0.011} & 0.149 & 0.197 & 0.193 & 0.010& 0.164& 0.239& 0.181 & 0.01& 0.145& 0.220 \\ 
    Arm circumference       & 0.283 & \textbf{0.052} & 0.165 & 0.409 & 0.358 & 0.035& 0.246& 0.490& 0.306 & 0.031& 0.216& 0.435 \\ 
    Arm length              & 0.329 & \textbf{0.029} & 0.270 & 0.395 & 0.364 & 0.018& 0.298& 0.423& 0.334 & 0.017& 0.271& 0.398 \\ 
    Forearm circumference   & 0.270 & \textbf{0.038} & 0.184 & 0.367 & 0.310 & 0.022& 0.233& 0.402& 0.264 & 0.019& 0.200& 0.342 \\
    Forearm length          & 0.254 & \textbf{0.027} & 0.197 & 0.300 & 0.268 & 0.015& 0.216& 0.328& 0.241 & 0.015& 0.169& 0.297 \\ 
    Thigh circumference     & 0.551 & \textbf{0.073} & 0.395 & 0.733 & 0.625 & 0.059& 0.412& 0.843& 0.616 & 0.056& 0.448& 0.870 \\
    Thigh length            & 0.512 & \textbf{0.046} & 0.415 & 0.590 & 0.409 & 0.028& 0.298& 0.520& 0.379 & 0.024& 0.304& 0.474 \\ 
    Calf circumference      & 0.354 & \textbf{0.042} & 0.248 & 0.458 & 0.392 & 0.030& 0.266& 0.523& 0.373 & 0.029& 0.282& 0.482 \\
    Calf length             & 0.424 & \textbf{0.040} & 0.348 & 0.504 & 0.419 & 0.025& 0.344& 0.519& 0.403 & 0.026& 0.294& 0.506 \\ 
    Foot width              & 0.098 & \textbf{0.007} & 0.083 & 0.115 & 0.102 & 0.005& 0.079& 0.126& 0.093 & 0.005& 0.077& 0.108 \\ 
    Heel to ball length     & 0.198 & \textbf{0.014} & 0.169 & 0.225 & 0.201 & 0.011& 0.156& 0.245& 0.182 & 0.009& 0.151& 0.216 \\ 
    Heel to toe length      & 0.269 & \textbf{0.019} & 0.229 & 0.306 & 0.271 & 0.013& 0.216& 0.323& 0.246 & 0.012& 0.198& 0.286\\ \hline
    \end{tabular}
    }
  \caption{Statistics of the anthropometric measures (in mm) on the 100k dataset compared with real measurements collected from~\cite{gordon20142012}. The minimum, maximum values along with the mean ${\mu}$ and the standard deviation ${\sigma}$ are reported.}
  \label{table:statistics-anthro}
\end{table*}


%% file: 09_03_anthro_regressor.tex
\subsection{Details of the Anthropometric regressor}
\label{Section:details-antrho-reg}

\begin{table}[!h]
\centering{
\begin{tabular}{|l|c|}
\hline
 & MSE error [mm] \\
\hline
no mask & 0.086 \\
binary mask & 0.0011 \\
soft exponential & 0.000142 \\
one model & 0.0010 \\
\hline
\end{tabular}
}
\caption{Mean Euclidean distance on the test dataset for the Anthro regressor model for different masks.}
\label{tab:masks}
\end{table}

In order to facilitate the evaluation of our system on the Humans in the Wild dataset~\cite{Shapy:CVPR:2022}, we trained a regression model that extracts the anthropometric measurements from registered meshes. 

\begin{figure*}[h]
  \centering
  \begin{subfigure}[t]{\textwidth}
      \includegraphics[bb=0 0 7200 2400, width=\linewidth, trim={17cm 18cm 4cm 16cm}, clip]{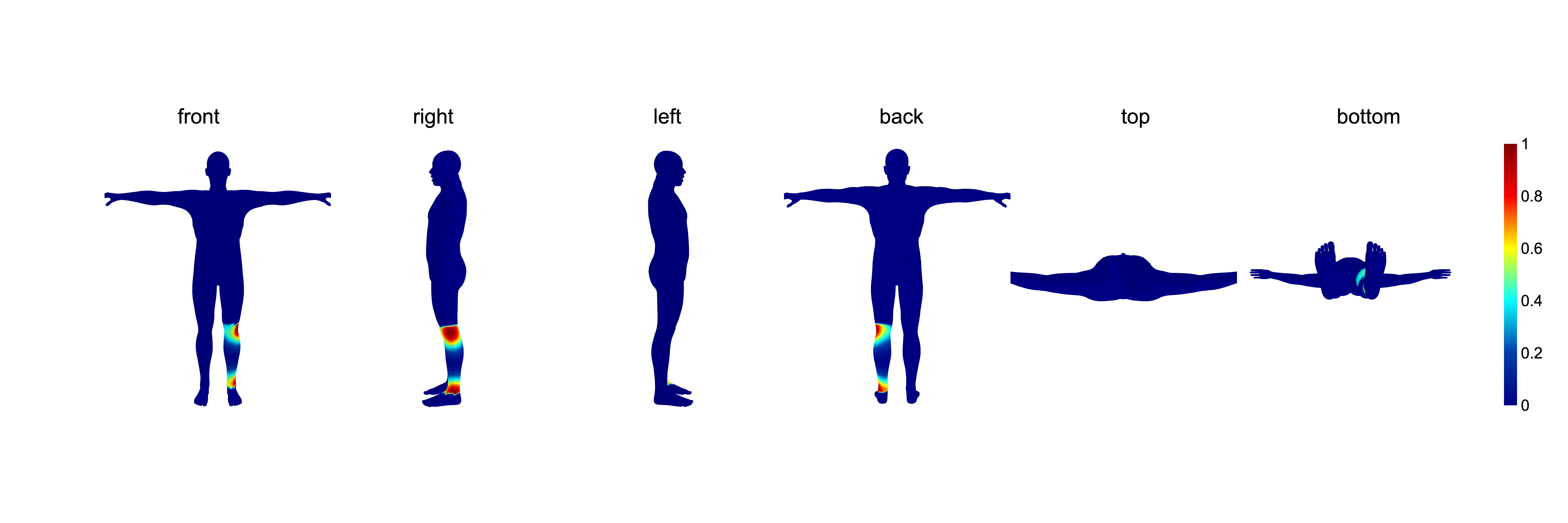}
      \caption{Left calf length male soft exponential mask}
        \label{fig:buffers-mask-soft-calf}
    \end{subfigure}
    \hspace{1.5em}
    \begin{subfigure}[t]{\textwidth}
\includegraphics[bb=0 0 7200 2400, width=\linewidth, trim={17cm 18cm 4cm 16cm}, clip]{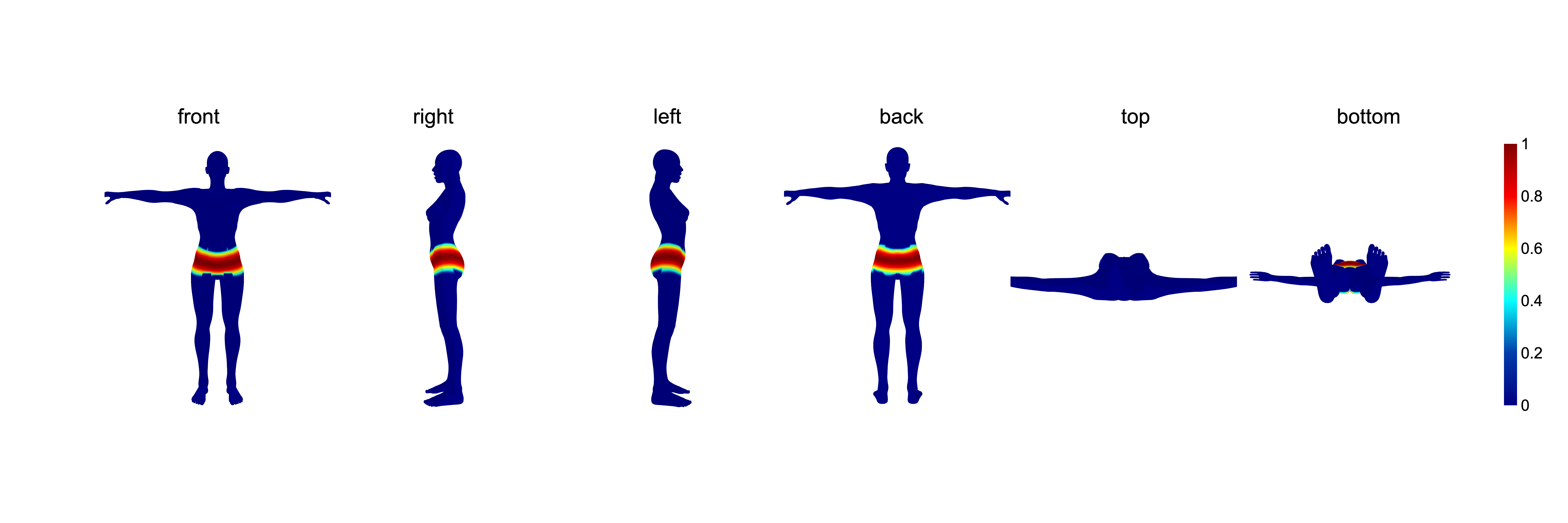}
        \caption{Hip circumference female soft exponential mask}
        \label{fig:buffers-mask-soft-hip}
    \end{subfigure}
    \hspace{1.5em}
    \begin{subfigure}[t]{\textwidth}
\includegraphics[bb=0 0 7200 2400, width=\linewidth, trim={17cm 18cm 4cm 16cm}, clip]{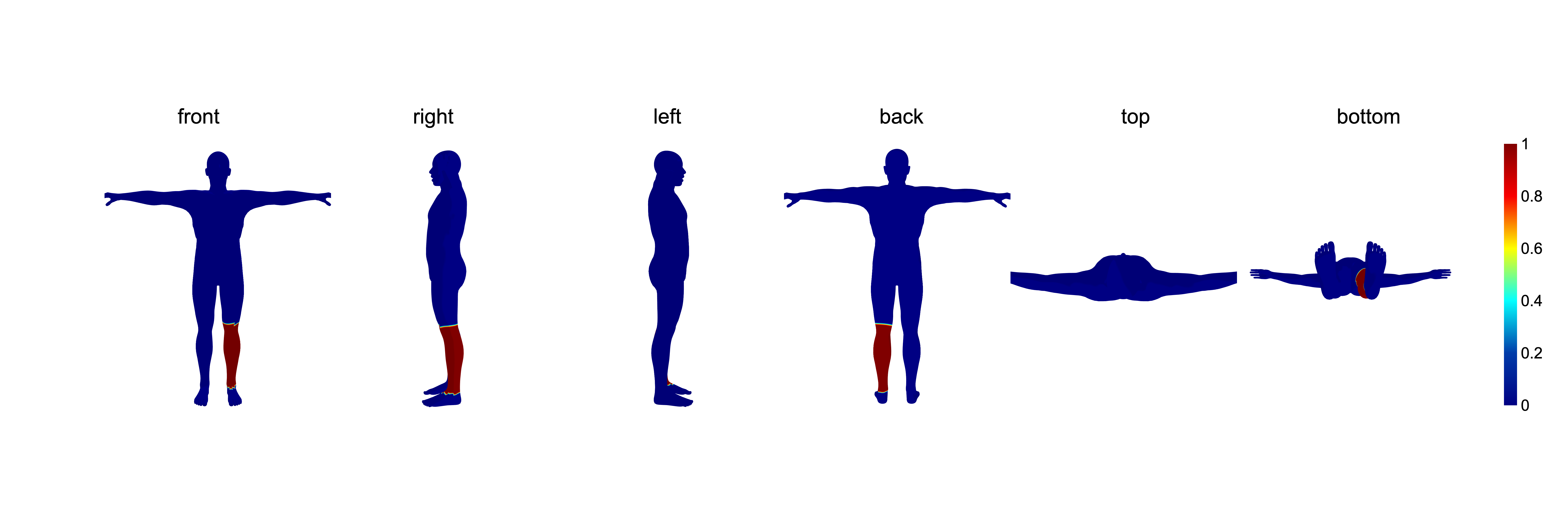}
        \caption{Left calf length male binary mask}
        \label{fig:buffers-mask-hard-calf}
    \end{subfigure}
    \hspace{1.5em}
    \begin{subfigure}[t]{\textwidth}
\includegraphics[bb=0 0 7200 2400, width=\linewidth, trim={17cm 18cm 4cm 16cm}, clip]{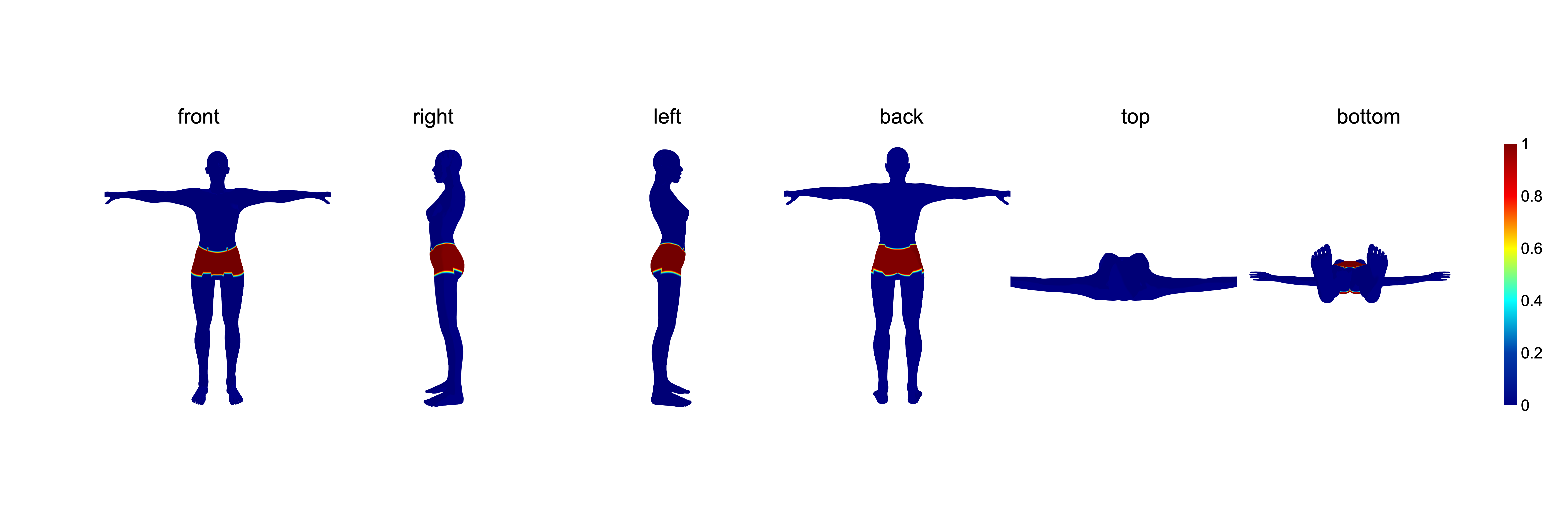}
        \caption{Hip circumference female binary mask}
        \label{fig:buffers-mask-hard-hip}
    \end{subfigure}
  \caption{After isolating the mesh related volume, we calculated the minimal Euclidean distance $d_\text{min}$ between each vertices of the volume and the ones used for the anthropometric measure. For the soft exponential mask (Figs.~\ref{fig:buffers-mask-soft-calf} and~\ref{fig:buffers-mask-soft-hip}), the final weight in the mask associated to the vertex point is $e^{-d_\text{min}}$. For the binary mask (Figs.~\ref{fig:buffers-mask-hard-calf} and~\ref{fig:buffers-mask-hard-hip}), we binarized the distance with a threshold equal to 0.5.
}
  \label{fig:masks}
\end{figure*}

For each of the 36 anthropometric measurements defined in sec. 5.1 in core paper, we trained an expert Multi Layer Perceptron~\cite{haykin1994neural} comprised of three layers. To ensure the disentanglement between the anthropometric measures, we designed a mask for each measure to isolate the vertices on the mesh related to the measure as shown in Fig.~\ref{fig:masks}.
This design prevents interference from other vertices in the prediction of a measure on a specific body part. 
During training, for each measurement the input mesh is masked with a specific mask, as depicted in Fig.~\ref{fig:masks}. 
At inference time, we aggregate the outputs of all the 36 models.

We conducted different experiments corresponding to specialized models for each anthropometric measurement using no mask or a binary mask or a soft exponential mask (described in the Fig.~\ref{fig:masks}) but also one general model which predicts all the anthropometric measurements at once. The results are reported in the Tab.~\ref{tab:masks}.
The specialized models using soft exponential masks have proven the best performance, thus they were chosen as reference. 

%% file: 09_04_mesh_skinner_rigger_arch.tex
\subsection{Deep Skinner architecture}
\label{Section:details-skinner}
In this section, we describe the design details of the deep neural network responsible for the rigging, skinning and posing of the meshes.
The goal of this network is to estimate the pose correctives $\Delta \Tilde{\mathcal{X}}^A_{\bar{\theta}}$ relative to the generated mesh in bind pose $\Tilde{\mathcal{X}}^A_b$ and the desired pose $\bar{\theta}$.
The forward pass of this network is twofold:
\begin{enumerate}[i.]
    \item an encoder concatenates the bind pose vertices $\mathcal{X}^A_b$ and the desired pose $\bar{\theta}$ (after Fourier encoding), and produces a feature vector $h$;
    \item a decoder reconstructs the offsets $\Delta \Tilde{\mathcal{X}}^A_{\bar{\theta}}$ from $h$.
\end{enumerate}

The encoder shares the same core block of the generative network, composed of a fully connected (FC) layer followed by a batch normalization (BN) layer and a parametric rectified linear unit (PReLU) activation function.
The decoding stage is made of a FC layer with 128 input features and 128 output features, and a FC layer with 128 input features and $3V$ output features. The latter passes through a PReLU activation function to produce the offsets estimate.

We performed ablation studies to determine the best design, in terms of mesh reconstruction ability.~\cref{tab:ablation_skinner} reports different combinations of architectures and number of Fourier encoding components, along with the mean vertex-to-vertex distance on the test dataset.
From our experiments, we can state that a simpler network achieves better results than deeper networks.
Moreover, the batch normalization layers are required to drive the optimization towards the desired solutions.

\begin{table}[]
\resizebox{\columnwidth}{!}{
\begin{tabular}{|cccc|c|}
\hline
encoding blocks & decoding FC layers & BN & Frequencies & P2P \\
\hline
1 & 2 & yes  & 32    & 4  \\
1 & 2 & yes  & 16    & 4  \\
3 & 2 & yes  & 16    & 15 \\
1 & 1 & no   & 16    & 29 \\
1 & 2 & yes  & 0     & 56 \\

\hline
\end{tabular}
}
\caption{Mean vertex-to-vertex distance (in mm) on held out test dataset for the Skinner model with different architecture designs, and number of Fourier components.}
\label{tab:ablation_skinner}
\end{table}

%% file: 09_06_extended_results.tex
\subsection{Random Fourier Encoding}
Tab.~\ref{tab:ablations_fourier} presents models with varying levels of Fourier Frequencies used for encoding of conditionals. 
By visual inspecting the training and test samples, we concluded that Fourier-encoding helped in preventing overfitting both the generator and skinner networks.

\begin{table}[h]
\centering
\resizebox{\linewidth}{!}{
\begin{tabular}{|l|l|l|l|l|l|l|l|}
\hline
Frequencies & P2P  & chest c. & hip c. & arm c. & calf c. & thigh c. & waist c. \\ \hline
8                   & \textbf{3.83} & \textbf{46.9}      & \textbf{9.79}     & \textbf{6.22}     & \textbf{4.53}      & 9.93       & \textbf{12.3}       \\ \hline
16                  & 5.04     &    60.4         &   14.9        &   14.5        &   9.60         &    \textbf{7.08}          &   21.6          \\ \hline
32                  &  4.87    &  57.7           &   13.1        &  12.4         &  8.40          & 7.93            &  18.0           \\ \hline
\end{tabular}
}
\caption{\textbf{Effect of Fourier Frequencies used in encoding} P2P loss is the mean vertex-to-vertex  distance in mm on held out test dataset. Similarly chest, hip, arm, calf, thigh, waist represents error in measurements of the circumferences in mm.}
\label{tab:ablations_fourier}
\end{table}


\subsection{Generative Performance}
This section details the generation of qualitative results presented Fig.~4~(a, b, c) in the core paper.
For each sub figure, two human measurement vectors $A1, A2$ are samples such that they posses desired physical attributes. In the case of Fig. 4a, $A1$ corresponds to dense measurements of a female with lean body, and $A2$ corresponds to female with the similar height but large circumferential measurements. Tab.~\ref{tab:interpolation_table} presents the $(A1, A2)$ pairs for the figures a, b, c respectively. Then, three intermediate measurements between $A1, A2$ are obtained by linear interpolation.
It can be seen that model is able to generate smooth interpolation in mesh space that respects the prescribed measurements. 

\begin{table}[t]
\centering
\resizebox{\linewidth}{!}{
\begin{tabular}{|l|cc|cc|cc|}
\hline
\textbf{Figure}                  & \multicolumn{2}{c|}{\textbf{Fig. 4 a}}                         & \multicolumn{2}{c|}{\textbf{Fig. 4 b}} & \multicolumn{2}{c|}{\textbf{Fig. 4 c}} \\ \hline \hline
Notation                & A1                                    & A2            & A1            & A2            & A1            & A2            \\ \hline
sex                     & F                                     & F             & M             & M             & \textbf{M}    & \textbf{F}    \\ \hline
waist circumference     & \textbf{0.61}                         & \textbf{1.13} & \textbf{0.91} & \textbf{0.75} & 0.73          & 0.73          \\ \hline
chest circumference     & \textbf{0.90}                         & \textbf{1.36} & \textbf{1.11} & \textbf{0.94} & 0.93          & 1.07          \\ \hline
hip circumference       & \textbf{0.85} & \textbf{1.22} & \textbf{1.01} & \textbf{0.85} & \textbf{0.84} & \textbf{1.00} \\ \hline
height                  & 1.60                                  & 1.62          & \textbf{1.99} & \textbf{1.50} & 1.97          & 1.98          \\ \hline
shoulder width          & 0.27                                  & 0.28          & \textbf{0.31} & \textbf{0.26} & 0.30          & 0.32          \\ \hline
torso height from back  & 0.34                                  & 0.35          & \textbf{0.41} & \textbf{0.31} & 0.42          & 0.42          \\ \hline
torso height from front & 0.29                                  & 0.32          & \textbf{0.35} & \textbf{0.26} & 0.34          & 0.36          \\ \hline
head circumference      & 0.53                                  & 0.54          & 0.55          & 0.53          & 0.55          & 0.55          \\ \hline
neck circumference      & \textbf{0.37} & \textbf{0.47} & \textbf{0.48} & \textbf{0.40} & 0.42          & 0.45          \\ \hline
head height             & 0.24                                  & 0.24          & 0.25          & 0.24          & 0.25          & 0.25          \\ \hline
midline neck length     & \textbf{0.09} & \textbf{0.07} & \textbf{0.10} & \textbf{0.08} & 0.11          & 0.11          \\ \hline
lateral neck length     & 0.11                                  & 0.10          & \textbf{0.13} & \textbf{0.10} & 0.14          & 0.14          \\ \hline
hand size               & 0.16                                  & 0.17          & \textbf{0.19} & \textbf{0.16} & 0.19          & 0.19          \\ \hline
arm circumference       & \textbf{0.23} & \textbf{0.34} & \textbf{0.33} & \textbf{0.27} & 0.26          & 0.26          \\ \hline
arm length              & 0.29                                  & 0.30          & \textbf{0.39} & \textbf{0.29} & 0.38          & 0.36          \\ \hline
forearm circumference   & \textbf{0.22} & \textbf{0.33} & \textbf{0.29} & \textbf{0.24} & 0.25          & 0.26          \\ \hline
forearm length          & 0.23                                  & 0.23          & \textbf{0.30} & \textbf{0.20} & 0.30          & 0.29          \\ \hline
thigh circumference     & \textbf{0.48} & \textbf{0.63} & \textbf{0.59} & \textbf{0.50} & 0.50          & 0.57          \\ \hline
thigh length            & 0.48                                  & 0.47          & \textbf{0.58} & \textbf{0.42} & 0.57          & 0.59          \\ \hline
calf circumference      & \textbf{0.31} & \textbf{0.38} & 0.38          & 0.33          & 0.33          & 0.37          \\ \hline
calf length             & 0.38                                  & 0.38          & \textbf{0.50} & \textbf{0.37} & 0.49          & 0.49          \\ \hline
foot width              & 0.09                                  & 0.09          & \textbf{0.11} & \textbf{0.08} & 0.10          & 0.11          \\ \hline
heel to ball length     & 0.18                                  & 0.18          & \textbf{0.22} & \textbf{0.17} & 0.21          & 0.22          \\ \hline
heel to toe length      & 0.25                                  & 0.25          & \textbf{0.29} & \textbf{0.23} & 0.29          & 0.30          \\ \hline
\end{tabular}
}
\caption{\textbf{Measurement data (in mm) for generation of Figure 4}. The measurement pairs $(A1, A2)$ for the figures 4.a, 4.b, 4.c are presented respectively. If there exists a significant difference between certain measurement, it is highlighted in bold. For example, in Fig. 4c, $A1, A2$ pair corresponds to male and female with similar physical characteristics that only differs in terms of gender and hip circumference.}
\label{tab:interpolation_table}
\end{table}

%% file: 09_07_b2a.tex
\subsection{Regression from SMPLX parameters}\label{supp-subsec:b2a}

\begin{figure*}
    \centering
    \includegraphics[bb=0 0 577 141, width=\textwidth]{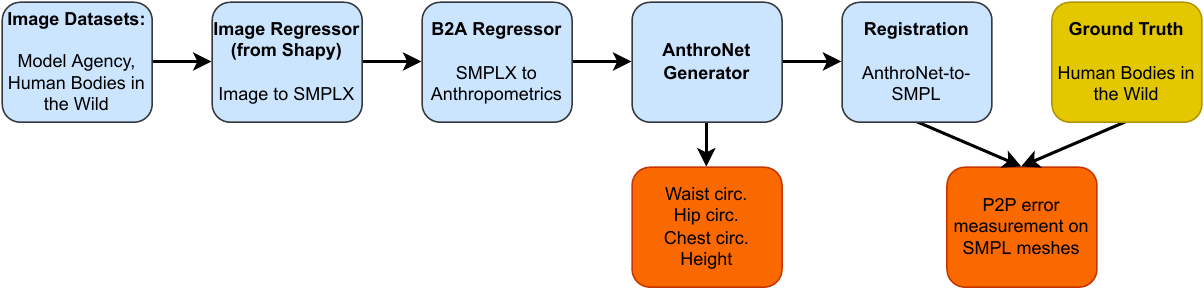}
    \caption{The proposed pipeline to generate AnthroNet meshes by regressing the set of dense anthropometric measures from SMPLX shape parameters ($\beta$). For evaluation, the AnthroNet mesh is registered to SMPL mesh, and then it is compared with the ground truth mesh.}
    \label{fig:b2a_pipeline}
\end{figure*}

In this section, we describe a regression module, referred to as Beta-to-Anthropometrics (B2A), which allows us to regress the set of dense anthropometric measures from SMPL-X shape parameters ($\beta$).
In other words, this module allows us to use any pretrained image regression module that provides shape parameters, either in SMPL or SMPL-X formats, to generate AnthroNet meshes. Due to the numerous capabilities of AnthroNet meshes in downstream tasks, e.g., dense anthropometry, rigging, skinning, and posing, the B2A module bridges the gap between any human body model in SMPL or SMPL-X formats and the AnthroNet system. As shown in Fig.~\ref{fig:b2a_pipeline}, B2A regresses the set of anthropometric measures required by our generative model from SMPL-X shape parameters. 

To train B2A, we employed the image-regression module from Shapy framework to obtain SMPL-X shape parameters of the subjects in the Model Agency dataset. Then, AnthroNet equivalent of these SMPL-X meshes are obtained through the SMPLX-to-AnthroNet registration pipeline, which allows us to compute the set of dense anthropometrics for each subject in the Model Agency dataset. This process provides us the training data for the B2A module ($B2A(): \beta \xrightarrow[]{} A$). B2A is a three-layer MLP, which uses mean squared error as the loss function and Adam optimizer as the optimization tool. To train the B2A module, we investigated the effect of Fourier encoding of anthropometric measures on the accuracy of regression. As the results in Tab.~\ref{tab:B2A_training} show, the best results are obtained when no Fourier encoding is used. 

\begin{table}[]
    \centering
    \resizebox{\columnwidth}{!}{
    \begin{tabular}{|c|c|c|c|}
    \hline 
    \textbf{Frequencies}  &  \textbf{Training loss}  & \textbf{Validation loss} & \textbf{Test loss}\\
    \hline 
    0  & 0.015 & 0.014 & 0.011 \\
    8  & 0.016 & 0.020 & 0.016 \\
    32 & 0.014 & 0.022 & 0.020\\
    \hline
    \end{tabular}
    }
    \caption{\textbf{Effect of Fourier encoding in training of B2A module} Please note that loss is the mean squared error between the predictions and the target anthropometric measures. It is worth highlighting that the training, validation, and test sets in this table are different splits of the Model Agency dataset. }
    \label{tab:B2A_training}
\end{table}

To evaluate the efficacy of B2A module in translating SMPL(-X) mesh system to AnthroNet by generating meshes with high fidelity to their SMPL(-X) counterparts, we deploy the Human Bodies in the Wild (HBW) dataset. As shown in Fig.~\ref{fig:b2a_pipeline}, the process here is to go from SMPL-X shape parameters (obtained from Shapy image regressor) to dense anthropometric measurements and then to AnthroNet mesh. Finally, the AnthroNet mesh is registered to SMPL format to calculate the P2P error with the ground truth mesh. This pipeline not only evaluates the efficacy of the B2A module but also assesses the fidelity of both AnthroNet generator and the AnthroNet-to-SMPL registration pipeline. The results of this evaluation are reported in Tab.~\ref{tab:appendix_image_regressor_compare}. It is worth highlighting that due to the unavailability of ground truth meshes for the test set of HBW, we have evaluated the AnthroNet on the validation set of HBW. Please note that HBW dataset has never been used in any development phases of AnthroNet, and thus, the validation split of HBW is indeed as unseen data for AnthroNet. 

Due to the fact that we are using the image-regression module from Shapy framework to regress shape parameters from images, it is expected that our performance is lower-bounded by the performance of the Shapy framework. However, given the fact that our pipeline encompasses several modules that are trained on different datasets and operate based non-deterministic functions such as optimization procedures, the reported performance for AnthroNet in Tab.~\ref{tab:appendix_image_regressor_compare} suggests a high degree of fidelity to the predictions of the image-regressor and a high degree of robustness/generalization in its entire pipeline. In other words, these results suggest that AnthroNet can be coupled with any SMPL(-X)-based image-regression module while imposing minimal loss on the reconstructed meshes (compared to ground truth) and providing several downstream benefits through the AnthroNet system. 

\begin{table}[]
    \centering
    \resizebox{\columnwidth}{!}{%
    \begin{tabular}{|l|c|c|c|c|c|c|}
    \hline
    Method                                          & Model  &  Height  &  Chest &  Waist  &  Hips &  P2P \\
    \hline
    SMPLR~\cite{SMPLR}                              & SMPL   & 182 & 267 & 309 & 305 & 69 \\
    STRAPS~\cite{STRAPS}                            & SMPL   & 135 & 167 & 145 & 102 & 47 \\
    SPIN~\cite{SPIN}                                & SMPL   & 59  & 92  & 78  & 101 & 29 \\
    TUCH~\cite{TUCH}                                & SMPL   & 58  & 89  & 75  & 57  & 26 \\
    Sengupta et al.~\cite{sengupta2021hierarchical} & SMPL   & 82  & 133 & 107 & 63  & 32 \\
    ExPose~\cite{EXPOSE}                            & SMPL-X & 85  & 99  & 92  & 94  & 35 \\
    SHAPY~\cite{Shapy:CVPR:2022}                    & SMPL-X & 51  & 65  & 69  & 57  & 21 \\
    AnthroNet (ours)                                & SMPL-X & 86  & 79  & 90  & 86 & 33 \\
    \hline 
    \end{tabular}
    }
    \caption{\textbf{Performance comparison of AnthroNet when coupled with a pretrained image regressor.} Values are the error in millimeters and evaluated over the HBW dataset~\cite{Shapy:CVPR:2022}. Please note that due to the absence of ground truth, AnthroNet is evaluated on the validation split of the HBW dataset while its counterparts are evaluated on the test split.}
    \label{tab:appendix_image_regressor_compare}
\end{table}